\documentclass[11pt]{article}

\usepackage{natbib}
\usepackage{amsmath}
\usepackage[dvips]{graphicx}
\usepackage{url}
\usepackage{color}

\newif\ifsubmit
\submittrue

\newenvironment{eqnnon}{\begin{equation}}{\end{equation}}
\newenvironment{eqnarraynon}{\begin{eqnarray}}{\end{eqnarray}}

\newcommand{\correction}{}

\begin{document}

\title{Accelerating Kernel Classifiers Through Borders Mapping}

\author{Peter Mills\\\textit{peteymills@hotmail.com}}

\maketitle

\ifsubmit
\newcommand{\decision}{g}
\newcommand{\lindecision}{\decision}
\newcommand{\bordernormal}{\vec v}
\newcommand{\borderconst}{b}

\newcommand{\vectorkernel}{K}		
\newcommand{\scalarkernel}{K}		
\newcommand{\kernelparam}{\vec \kappa}	
\newcommand{\norm}{N}			
\newcommand{\nsample}{n}		
\newcommand{\dimension}{D}		
\newcommand{\densityestimator}{\tilde p}
\newcommand{\bandwidth}{\sigma}
\newcommand{\coord}{x}
\newcommand{\samplecoord}{x}
\newcommand{\sample}{\vec \samplecoord}
\newcommand{\testcoord}{x}
\newcommand{\testpoint}{\vec \testcoord}
\newcommand{\point}{\vec \coord}
\newcommand{\classlabel}{y}
\newcommand{\class}{c}
\newcommand{\probability}{P}
\newcommand{\condprob}{p}
\newcommand{\jointprob}{p}
\newcommand{\kernelsum}{W}

\newcommand{\svmcoeff}{w}
\newcommand{\expandedspace}{\vec\phi}
\newcommand{\svmdecision}{\decision_{svm}}
\newcommand{\svmbordernormal}{\bordernormal}
\newcommand{\svmborderconst}{\borderconst}
\newcommand{\svmcost}{C}

\newcommand{\decisionfunction}{\tilde r}
\newcommand{\diffcondprob}{r}
\newcommand{\kerneldecision}{\decisionfunction_{kern}}
\newcommand{\vbdecision}{\decisionfunction_{vb}}
\newcommand{\bordervector}{\vec b}
\newcommand{\nborder}{n_b}
\newcommand{\borderdecision}{\decision_{border}}
\newcommand{\distance}{d}		
\newcommand{\scalarkernelderiv}{\dot \scalarkernel}

\newcommand{\svmprob}{\decisionfunction_{svm}}
\newcommand{\svmprobcoeffA}{A}
\newcommand{\svmprobcoeffB}{B}
\newcommand{\borderprob}{\decisionfunction_{border}}

\newcommand{\nclass}{n_c}	
\newcommand{\multiprob}{p}	
\newcommand{\lmult}{\lambda}	

\newcommand{\featurevolume}{V}	
\newcommand{\mintrain}{n_0}	
\newcommand{\minborders}{n_{b0}}	

\newcommand{\rbkernelparam}{\gamma}

\newcommand{\datafraction}{f}

\newcommand{\confusion}{\eta}		
\newcommand{\ntest}{n_{test}}		
\newcommand{\entropy}{H}
\newcommand{\priorentropy}{\entropy_i}
\newcommand{\posteriorentropy}{\entropy(i|j)}
\newcommand{\accuracy}{a}
\newcommand{\UC}{U(i|j)}

\newcommand{\classsize}{n}		
\newcommand{\subsample}{\alpha}		
\newcommand{\submultcoef}{C}
\newcommand{\subexp}{\zeta}

\newcommand{\dataset}[1]{\textsf{#1}}

\else
  
\fi

\begin{abstract}
  Support vector machines (SVM) and other kernel techniques represent a family of powerful statistical 
classification methods with high accuracy and broad applicability.
Because they use all or a significant portion of the training data, 
however, they can be slow, especially for large problems.
Piecewise linear classifiers are similarly versatile, yet have the additional
advantages of simplicity, ease of interpretation and, if the number of 
component linear classifiers is not too large, speed.
Here we show how a simple, piecewise linear classifier can be trained from
a kernel-based classifier in order to improve the classification speed.
The method works by finding the root of the difference in conditional
probabilities between pairs of opposite classes to build up a representation
of the decision boundary.
When tested on 17 different datasets, it succeeded in improving the
classification speed of a SVM for 12 of them by up to two orders-of-magnitude.
Of these, two were less accurate than a simple, linear classifier.
The method is best suited to problems with continuum features
data and smooth probability functions.
Because the component linear classifiers are built up individually from an
existing classifier, rather than through a simultaneous optimization procedure,
the classifier is also fast to train.

\end{abstract}

\subsection*{Keywords}
\textbf{class borders,
multi-dimensional root-finding,
binary classifier,
multi-class classifier,
kernel methods,
non-parametric statistics} 

\newpage

\tableofcontents




\section{Introduction}

Linear classifiers are well studied in the literature. Methods such as
the perceptron, Fisher discriminant, logistic regression and now linear
{\it support vector machines} (SVMs) \citep{Michie_etal1994}
are often appropriate for relatively simple,
{\it binary classification} problems in which both classes are closely 
clustered or otherwise well separated.
An obvious extension for more complex problems is a {\it piecewise linear classifier} 
in which the {\it decision boundary} is built up from a series of linear classifiers.
Piecewise linear classifiers enjoyed some popularity during the early 
development of the field of machine learning 
\citep{Osborne1977,Sklansky_Michelotti1980,Lee_Richards1984,Lee_Richards1985}
and because of their versatility, generality and simplicity there has been recent renewed interest 
\citep{Bagirov2005,Kostin2006,Gai_Zhang2010,Webb2012,Wang_Saligrama2013,Pavlidis_etal2016}.

A {\it linear classifier} takes the form:
\begin{equation}
	\lindecision(\testpoint) = \bordernormal \cdot \testpoint + \borderconst
	\label{linear_classifier}
\end{equation}
where $\testpoint$ is a {\it test point} in the {\it feature} space,
$\bordernormal$ is a normal to the decision hyper-surface, 
$\borderconst$ determines the location of the decision boundary along the normal
and $\lindecision$ is the {\it decision function} 
which we use to estimate the class of the test point through its sign.

A piecewise linear classifier collects a set of such linear classifiers:
$\lbrace \bordernormal_i \rbrace = \lbrace \bordernormal_1, \bordernormal_2,
\bordernormal_3 ... \rbrace$; $\lbrace \borderconst_i \rbrace =
\lbrace \borderconst_1, \borderconst_2, \borderconst_3, ... \rbrace$.
The two challenges here are, first, how to efficiently train each of the
decision boundaries and, second, the related problem of how to partition the 
feature space to determine which linear decision boundary is used for a given 
test point.

In \citet{Bagirov2005} for instance, the decision function is defined by
dividing the set of linear classifiers and maximizing the minimum
linear decision value in each subset.
To train the classifier, a cost function is defined in terms
of this decision function and directly minimized using an
analog to the derivative for non-smooth functions \citep{Bagirov1999}.
Naturally, such an approach will be quite computationally costly,
especially for a large number of component linear classifiers.

Partitioning of the feature space can be separate from the discrimination 
borders \citep{Huang_etal2013} but more normally the discrimination borders 
are themselves sufficient to partition the feature space 
\citep{Osborne1977,Lee_Richards1984,Bagirov2005,Kostin2006}.
This means that all or a significant fraction of the component
linear classifiers must be evaluated.
In \citet{Kostin2006}, for instance, the linear classifiers form a decision
tree.

In the method described in this paper, the constant term, $\borderconst_i$,
is changed to a vector and the partitioning accomplished through a nearest
neighbours to this vector. 
Thus the zone of influence for each hyperplane
will be described by the Veronoi tessellation \citep{Kohonen2000}.
If the class domains are simply connected and don't curve back on themselves, 
then the partitions will also be shaped as hyper-pyramids, 
with the axes of the pyramids roughly perpendicular to the decision border.
A dot product with each of the vectors must be calculated, similar
to a linear classifier, but afterwards only a single linear decision function is
evaluated.

There seems to be some tension in the literature between training the
decision boundary through simultaneous optimization \citep{Bagirov2005,Wang_Saligrama2013} or through
methods that are more piece meal \citep{Gai_Zhang2010,Herman_Yeung1992,Kostin2006}.
Obviously, simultaneous optimization will be more accurate but also more computationally expensive.
In addition, finding global minima for cost functions 
involving more than a handful of hyper-surfaces will be all but impossible.
There is also the issue of separability. Many of the current 
methods seem designed with disjoint classes in mind \citep{Herman_Yeung1992}, for instance
\citet{Gai_Zhang2010}, who stick the hyper-plane borders between 
neighbouring pairs of opposite classes.
Yet there is no reason why a piecewise linear classifier cannot be just as
effective for overlapping classes.

The technique under discussion in this paper mitigates all of these issues 
because it is not a stand-alone method but requires estimates
of the conditional probabilities.
It is used to improve the time performance of kernel methods, or for that matter,
any binary classifier that returns a continuous decision function 
that can approximate a conditional probability.
This is done while maintaining, in all but a few cases, most of the accuracy.

Several of the piecewise linear techniques found in the literature work by
positioning each hyperplane between pairs of clusters or pairs of training
samples of opposite class 
\citep{Sklansky_Michelotti1980,Tenmoto_etal1998,Kostin2006,Gai_Zhang2010}.
Other unsupervised or semi-supervised classifiers work by placing the 
hyperplanes in regions of minimum density \citep{Pavlidis_etal2016}.
The method described in this paper in some senses combines these two techniques
by finding the root of the difference in conditional probabilities along a
line between two points of opposite class.
It will be tested on a kernel-based classifier called a support vector
machine (SVM) \citep{Michie_etal1994,Mueller_etal2001}
and evaluated based on how
well it improves classification speed and at what cost to accuracy.
It will also be compared to a simple, linear classifier as in (\ref{linear_classifier}).

Section \ref{theory} describes the theory of support
vector machines and linear classifiers
as well as the piecewise linear classifier or ``borders'' classifier
that will be trained on the SVM.
Section \ref{methods} describes the software and test datasets then in
section \ref{example} we analyze the different classification algorithms on
a simple, synthetic dataset.
Section \ref{results_section} outlines the results for 17 case studies
while in Section \ref{discussion} we discuss the results.
Section \ref{conclusion} concludes the paper.

\section{Theory}

\label{theory}

We are interested in training a binary classification model based on a set of 
{\it training data}, $\lbrace \sample_i: \classlabel_i|~i \in [1,\nsample] \rbrace$, where $\sample_i$
is a vector of features data and 
$\classlabel_i \in \lbrace -1,~ +1 \rbrace$ is the corresponding
{\it class label}.
In this section we will describe methods for training the three types of
models tested in this paper: a basic linear classifier, a support vector
machine (SVM) and a borders classifier.

\subsection{Support vector machines}

Even though different software is used for each, 
we describe support vector machines (SVMs) and linear classifiers in the same
section since a linear classifier is just a restricted version of a SVM.
The decision function for a SVM is defined as follows:
\begin{equation}
	\svmdecision(\testpoint) = \sum_{i=1}^\nsample \svmcoeff_i \classlabel_i \vectorkernel (\testpoint, \sample_i) + \svmborderconst
	\label{svm_decision}
\end{equation}
where $\vectorkernel$ is a {\it kernel function}, $\svmborderconst$ is a
constant, $\lbrace \svmcoeff_i \rbrace$ are a sparse set of coefficients
and $\nsample$ is the number of training samples.
The class is determined as in (\ref{linear_classifier}) by the sign of the decision value:
\begin{equation}
	\class(\testpoint) = \frac{\svmdecision(\testpoint)}{|\svmdecision(\testpoint)|}
	\label{class_value}
\end{equation}
where for convenience, the class labels are defined as
$c \in \lbrace -1, 1 \rbrace$.

The constant and coefficients are fitted by solving the following, quadratic
optimization problem:
\begin{equation}
	\max_{\lbrace \svmcoeff_i \rbrace} \left [ \sum_i \svmcoeff_i 
	- \frac{1}{2} \sum_{i, j} \svmcoeff_i \svmcoeff_j \classlabel_i \classlabel_j \vectorkernel(\sample_i, \sample_j) \right ] \label{dual_problem}
\end{equation}
subject to the following constraints:
\begin{eqnarray}
	0 \le \svmcoeff_i & \le & \svmcost \label{constraint1} \\
	\sum_i \svmcoeff_i \classlabel_i & = & 0 \label{constraint2}
\end{eqnarray}
where $\svmcost$ is the {\it cost} \citep{Mueller_etal2001,Chang_Lin2011}.
The coefficients thus derived are typically sparse and the training samples
for which $\svmcoeff_i$ are non-zero are called {\it support vectors}.

SVM models will be trained with LIBSVM \citep{Chang_Lin2011}.
LIBSVM has the option to return estimates of the conditional 
probabilities by transforming the raw decision function using logistic regression:
\begin{equation}
	\svmprob(\testpoint) = \tanh \left (\frac{\svmprobcoeffA \svmdecision(\testpoint)+ \svmprobcoeffB}{2} \right )
	\label{svm_prob}
\end{equation}
where $\svmprobcoeffA$ and $\svmprobcoeffB$ are coefficients derived from
the training data via a nonlinear fitting technique \citep{Platt1999,Lin_etal2007} and $\svmprob$ is an estimator for the difference in conditional
probabilities--see Equation (\ref{rdef}), below.
Probability estimates are needed for the borders training: see the following
section.

In a linear SVM, the kernel function, $\vectorkernel$, is a simple dot product:
\begin{eqnnon}
	\svmdecision(\testpoint) = \sum_i \svmcoeff_i \classlabel_i \testpoint \cdot \sample_i + \svmborderconst
\end{eqnnon}
By exchanging the order of the summation operators, 
we can show that the normal to the
decision hyper-plane, $\bordernormal$, takes on the following value:
\begin{eqnnon}
	\bordernormal = \sum_i \svmcoeff_i \classlabel_i \sample_i
\end{eqnnon}
with which classifications may be performed using the simpler formula in 
(\ref{linear_classifier}).

\subsection{Borders classification}

\label{border_method}

In kernel SVM, the decision border exists only implicitly in a hypothetical,
abstract space. Even in linear SVM, if the software is generalized to 
recognize the simple dot product as only one among many possible kernels,
then the decision function may be built up, as in (\ref{svm_decision})
through a sum of weighted kernels. This is the case for LIBSVM.
The advantage of an explicit decision border as in 
(\ref{linear_classifier}) is that it is fast. 
The problem with a linear border is that, except for a
small class of problems, it is not very accurate.

In the binary classification method described in \citet{Mills2011},
a non-linear decision border is built up piece-wise from a collection of linear borders.
It is essentially a root-finding procedure for a decision function,
such as $\svmprob$ in (\ref{svm_prob}).
Let $\decisionfunction$ be a decision function
that approximates the difference in conditional probabilities:
\begin{equation}
	\decisionfunction(\point) \approx \diffcondprob(\point) = 
	\condprob(+1|\point) - \condprob(-1|\point)
	\label{rdef}
\end{equation}
where $\condprob(\class|\point)$ represents the conditional probabilities of
a binary classifier having labels $\class \in \lbrace -1, +1 \rbrace$.

The procedure is as follows: pick a pair of points on either side of the decision
boundary (the decision function has opposite signs). Good candidates are one
random training sample from each class. Then, zero the decision function
along the line between the points. This can be done as many times as needed
to build up a good representation of the decision boundary.
We now have a set of points, $\lbrace \bordervector_i \rbrace$, such that
$\decisionfunction(\bordervector_i)=0$ for every $i \in [1,~\nborder]$ where
$\nborder$ is the number of border samples.

Along with the border samples,  $\lbrace \bordervector_i \rbrace$, we also
collect a series of normal vectors, $\lbrace \bordernormal_i \rbrace$
such that:
\begin{eqnnon}
\bordernormal_i=\nabla_{\point}{\decisionfunction |_{\point=\bordervector_i}}
\end{eqnnon}
With this system, determining the class is a two step process.
First, the nearest border sample to the test point is found.
Second, we define a new decision function, $\borderdecision$, 
equivalent to (\ref{linear_classifier}), through a dot product with the normal:
\begin{eqnarray}
	i & = & \arg \min_j |\testpoint - \bordervector_j| \nonumber \\
	\borderdecision(\testpoint) & = & \bordernormal_i \cdot (\testpoint - \bordervector_i)
	\label{border_decision}
\end{eqnarray}
The class is determined by the sign of the decision function as in 
(\ref{class_value}).
The time complexity is independent of the number
of training samples, rather it is linearly proportional to the number of
border vectors, $\nborder$, a tunable parameter. The number required for
accurate classifications is dependent on the complexity of the decision
border.

The gradient of the SVM decision in (\ref{svm_prob}) is:
\begin{eqnnon}
	\nabla_{\point} {\svmprob} = \left [1 - \svmprob^2(\point) \right ] \sum_i \svmcoeff_i \classlabel_i \nabla_{\point} \vectorkernel(\point, \sample_i)
\end{eqnnon}

Gradients of the initial decision function are useful not only to derive normals to
the decision boundary, but also as an aid to root finding when searching for
border samples. If the decision function used to compute the border samples
represents an estimator for the
difference in conditional probabilities, then the raw decision value,
$\borderdecision$,
derived from the border sampling technique in (\ref{border_decision})
can also return estimates of the conditional probabilities with little
extra effort and little loss of accuracy, also using a sigmoid function:
\begin{equation}
	\borderprob(\testpoint) = \tanh \left [\borderdecision (\testpoint) \right ]
	\label{border_probability}
\end{equation}
This assumes that the class posterior probabilities,
$\condprob(\point | c)$, are approximately Gaussian near the border
\citep{Mills2011}.

The border classification algorithm returns an estimator,
$\borderprob$, for the difference in conditional probabilities of
a binary classifier using
equations (\ref{border_decision}) and (\ref{border_probability}).
It can be trained with the function in $\svmprob$ in (\ref{svm_prob}),
or any other 
continuous, differentiable, non-parametric estimator for the difference
in conditional probabilities, $\diffcondprob$.
At the cost of a small reduction in accuracy,
it has the potential to drastically reduce classification time for kernel
estimators and other non-parametric statistical classifiers,
especially for large training datasets,
since it has $O(\nborder)$ time complexity instead of $O(\nsample)$
complexity, where $\nborder$, the number of border samples, is a free parameter.
The actual number chosen
can trade off between speed and accuracy with rapidly diminishing returns
beyond a certain point. 
One hundred border samples ($\nborder=100$) is usually sufficient.
The computation of $\borderprob$ also involves very simple operations---
floating point addition, multiplication and numerical comparison, with no
transcendental functions except for the very last step (which can be omitted)---so the coefficient for the time complexity will be small.

A border classifier trained with SVM will be referred to as an
``SVM-borders'' classifier or an ``accelerated'' SVM classifier.
For more details on the algorithm, please refer to \citet{Mills2011},
in particular Sections 2.2, 2.3 and 3.3.

\subsection{Multi-class classification}

The border classification algorithm, like SVM, only works for binary 
classification problems. It is quite easy to generalize a binary classifier
to perform multi-class classifications by using several of them and the
number of ways of doing so grows exponentially with the number of classes.
Since LIBSVM uses the ``one-versus-one'' method \citep{Hsu_Lin2002} of 
multi-class classification, this is the one we will adopt. 

A major advantage of the
borders classifier is that it returns probability estimates.
These estimates have many uses including measuring the confidence of as well
as recalibrating the class estimates \citep{Mills2009,Mills2011}.
Thus the multi-class method
should also solve for the conditional probabilities in addition to returning
the class label.

In a one-vs.-one scheme, the multi-class conditional probabilities 
can be related to those of the binary classifiers as follows:
\begin{eqnnon}
	\diffcondprob_{ij}(\point) = \frac{\condprob(j|\point) - \condprob(i|\point)}{\condprob(i|\point) + \condprob(j|\point)}
	\label{bin2multi}
\end{eqnnon}
where $i\in[1,\nclass-1]$, $j\in[1,\nclass]$, 
$j>i$,
$\nclass$ is the number of classes, 
and $\diffcondprob_{ij}$ is the difference in conditional probabilities of
the binary classifier that discriminates between the $i$th and $j$th classes.
\citet{Wu_etal2004} transform this problem into the following linear system:
\begin{eqnarraynon}
	\multiprob_i \sum_{k|k \ne i} (\diffcondprob_{ki} + 1)^2 +
	\sum_{j|j \ne i} \multiprob_j (1 - r_{ij}^2) + \lmult & = & 0 \\ \nonumber
	\sum_j \multiprob_j & = & 1
	\label{multiclass}
\end{eqnarraynon}
where $\multiprob_i = \condprob(i | \point)$ is the $i$th multi-class 
conditional probability and $\lmult$ is a Lagrange multiplier.
They also show that the constraints not included in the problem, that
the probabilities are all positive, are always satisfied
and describe an algorithm for solving it iteratively, although a
simple matrix solver is sufficient.

\subsection{Skill scores}

\label{skill_scores}

It is important to evaluate a result based on skill scores that reliably reflect
how well a given classifier is doing.
Thus we will define the two scores used in this validation exercise 
since one in particular is not commonly seen in the literature even though it has several
attractive features.

Let $\lbrace \confusion_{ij} \rbrace$ be the confusion matrix, that is the number
test values for which the first classifier (the ``truth'') returns the $i$th class
while the second classifier (the estimate) returns the $j$th class.
Let $\ntest=\sum_i \sum_j \confusion_{ij}$ be the total number of test points.

The {\it accuracy} is given:
\begin{eqnnon}
\accuracy=\frac{\sum_i \confusion_{ii}}{\ntest}
\label{accuracy}
\end{eqnnon}
or simply the fraction of correct guesses.

The {\it uncertainty coefficient} is a more sophisticated measure based on the channel 
capacity \citep{Shannon}. It has the advantage over simple accuracy in that 
it is not affected by the relative size of each class distribution.
It is also not affected by consistent rearrangement of the class labels.

The entropy of the prior distribution is given:
\begin{eqnarraynon}
	\priorentropy & = & - \sum_i \left (\sum_j \frac{\confusion_{ij}}{\ntest} \right ) 
	\log \left (\sum_j \frac{\confusion_{ij}}{\ntest} \right )\\
	& = & - \frac{1}{\ntest} \left [\sum_i \left (\sum_j \confusion_{ij} \right ) 
	\log \left (\sum_j \confusion_{ij} \right )
	- \log \ntest \right ]
	\label{prior_entropy}
\end{eqnarraynon}
while the entropy of the posterior distribution is given:
\begin{eqnarraynon}
	\posteriorentropy & = & - \sum_i \sum_j \left ( \frac{\confusion_{ij}}{\ntest} \right ) \log \left (\frac{\confusion_{ij}}{\sum_i \confusion_{ij}} \right )
	\label{posterior_entropy} \\
	& = & - \frac{1}{\ntest} \left [ \sum_i \sum_j \confusion_{ij} \log \confusion_{ij} 
- \sum_j \left ( \sum_i \confusion_{ij} \right ) \log \left ( \sum_i \confusion_{ij} \right ) \right ]
\end{eqnarraynon}
The uncertainty coefficient is defined in terms of the prior entropy, $\priorentropy$, and the
posterior entropy, $\posteriorentropy$, as follows:
\begin{eqnnon}
	\UC = \frac{\priorentropy - \posteriorentropy}{\priorentropy}
	\label{uncertainty_coefficient}
\end{eqnnon}
and tells us: 
for a given test classification, how many bits of information 
on average does the estimate
supply of the true class value? \citep{Press_etal1992,Mills2011}.

\section{Software and data}

\label{methods}

\subsection{LIBLINEAR}

LIBLINEAR is a software package for linear statistical classification 
developed by the Machine Learning Group at the National Taiwan University.
It supports several different training methods including primal and dual
SVM as well as logistic regression.
The method that provided the most accurate results for each dataset was used.
The library may be downloaded at: 
\url{https://www.csie.ntu.edu.tw/~cjlin/liblinear} \citep{Fan_etal2008}.
LIBLINEAR is very fast for both training and prediction and it is especially
well suited to large-dimensional problems.

\subsection{LIBSVM}

LIBSVM is a machine learning software library for support vector machines 
developed by Chih-Chung Chang and Chih-Jen Lin of 
the National Taiwan University, Taipei, Taiwan \citep{Chang_Lin2011}.
It includes statistical classification using two regularization methods 
for minimizing over-fitting: 
{\it C-SVM} and {\it $\nu$-SVM}.
It also includes code for nonlinear regression and density estimation or
``one-class SVM''.
SVM models were trained using the \verb/svm-train/ command while
classifications done with \verb/svm-predict/.
LIBSVM can be found at: \url{https://www.csie.ntu.edu.tw/~cjlin/libsvm}.
LIBSVM is not very fast but tends to be very accurate for a wide variety
of problems.

\subsection{LibAGF}

Similar to LIBSVM, libAGF is a machine learning library but for variable kernel 
estimation \citep{Mills2011,Terrell_Scott1992} rather than SVM.
Like LIBSVM, it supports statistical classification, nonlinear regression
and density estimation.
It supports both Gaussian kernels and k-nearest neighbours.
It was written by Peter Mills and can be found at
\url{https://github.com/peteysoft/libmsci}.
To convert a LIBSVM model to a borders model,
the single command, \verb/svm_accelerate/, can be used.
Classifications are then performed with \verb/classify_m/.

\subsection{Datasets}

\label{datasets}

\begin{table}
  \caption{Summary of datasets used in the numerical trials.}
  \label{summary}
  \begin{tabular}{|l|llllll|}
	\hline
	& $\dimension$ & Type & $\nclass$ & Train & Test & Reference \\\hline
	heart & 13 & float & 2 & 270 & - & {\small \citep{Lichman2013}}\\
	shuttle & 9 & float & 7 & 43500 & 14500 & {\small \citep{King_etal1995}}\\
	sat & 36 & float & 6 & 4435 & 2000 & {\small \citep{King_etal1995}}\\
	segment & 19 & float & 7 & 2310 & - & {\small \citep{King_etal1995}} \\
	dna & 180 & binary & 3 & 2000 & 1186 & {\small \citep{Michie_etal1994}}\\
	splice &  60 & cat & 3 & 1000 & 2175 & {\small \citep{Michie_etal1994}}\\
	codrna &  8 & mixed & 2 & 59535 & 271617 & {\small \citep{Uzilov_etal2006}}\\
	letter &  16 & integer & 26 & 20000 & - & {\small \citep{Frey_Slate1991}}\\
	pendigits & 16 & integer & 10 & 7494 & 3498 & {\small \citep{Alimoglu1996}}\\
	usps & 256 & float & 10 & 7291 & 2001 & {\small \citep{Hull1994}}\\
	mnist & 665 & integer & 10 & 45000 & 10000 & {\small \citep{LeCun_etal1998}}\\
	ijcnn1 & 22 & float & 2 & 49990 & 91701 & {\small \citep{Feldkamp_Puskorius1998}}\\
	madelon & 500 & integer & 2 & 2000 & 600 & {\small \citep{Guyon_etal2004}}\\
	seismic & 50 & float & 2 & 78823 & 19705 & {\small \citep{Duarte_Hu2004}}\\
	mushrooms & 112 & binary & 2 & 8124 & - & {\small \citep{Iba_etal1988}}\\
	phishing & 68 & binary & 2 & 11055 & - & {\small \citep{Mohommad_etal2014}}\\
	humidity & 7 & float & 8 & 86400 & - & {\small \citep{Mills2009}}\\
	\hline
\end{tabular}

\end{table}

The borders classification algorithm was tested on a total of 
17 different datasets.
These will be briefly described in this section.
The collection covers a fairly broad range of size and types of problems, 
number of classes and number and types of attributes but with
the focus on larger datasets where the borders technique is actually useful.
Four of the datasets are from
the ``Statlog'' project \citep{Michie_etal1994,King_etal1995} 
and are nicknamed ``\dataset{heart}'', ``\dataset{shuttle}'', ``\dataset{sat}'' and ``\dataset{segment}''.
The heart disease (``\dataset{heart}'') dataset 
contains thirteen attributes of 270 patients along with one of two class labels denoting either the presence or absence of heart disease.
The dataset comes originally from the Cleveland Clinic Foundation and two versions are stored on the machine learning database of U. C. Irvine \citep{Lichman2013}.

The \dataset{shuttle} dataset is interesting because 
the classes have a very uneven distribution meaning that multi-class
classifiers
with a symmetric break-down of the classes, 
such as one-vs.-one,
tend to perform poorly.
The \dataset{shuttle} dataset comes originally
from NASA and was taken from an actual space shuttle flight.
The classes describe actions to be taken at different flight configurations.

The satellite (``\dataset{sat}'') dataset is a satellite remote-sensing land classification problem.
The attributes represent 3-by-3 segments of pixels in a Landsat 
image with the class corresponding to the type of land cover in the central pixel.
The segmentation (``\dataset{segment}'') dataset is also an image classification dataset consisting of 3-by-3
pixel sets from outdoor images.

The DNA dataset is concerned with classifying a 60 base-pair sequence of DNA into
one of three values: an intron-extron boundary, an extron-intron boundary or
neither of those two.
That is, during protein creation, part of the sequence is spliced out, with
the section kept being the intron and that spliced out being the extron.
There are two versions of it: one called ``\dataset{splice}'' with the original 
sequence of 4 nucleotide bases but only two classes 
and one called ``\dataset{dna}'' in which
the features data has been reprocessed so that
the 60 base values are transformed to 180 binary attributes but keeping the
original three classes \citep{Michie_etal1994}.
Another dataset from the field of microbiology is the ``\dataset{codrna}'' dataset
which deals with detection of non-coding RNA sequences
\citep{Uzilov_etal2006}.

There are four text-classification datasets: ``\dataset{letter}'', ``\dataset{pendigits}'',
``\dataset{usps}'' and ``\dataset{mnist}''.
The ``\dataset{letter}'' dataset is a text-recognition problem concerned with classifying
a character into one of the 26 letters of the alphabet based on processed
attributes of the isolated character \citep{Frey_Slate1991}.
The \dataset{pendigits} dataset is similar to the \dataset{letter} dataset except for
classifying numbers instead of letters \citep{Alimoglu1996}.
The ``\dataset{usps}'' dataset deals with classifying text for the purpose of mailing
letters \citep{Hull1994}.
The ``\dataset{mnist}'' dataset uses 28 by 28 pixel images to classify text into one
of ten different characters \citep{LeCun_etal1998}. 
Pixels that always take on the same value were removed.

Two of the datasets are machine-learning competition challenges.
The ``\dataset{ijcnn1}'' dataset is from the International Joint Conference on Neural
Networks Neural Networks Competition\citep{Feldkamp_Puskorius1998} while the ``\dataset{madelon}'' dataset comes 
from the International Conference on Neural Information Processing Systems
Feature Selection Challenge \citep{Guyon_etal2004}.

The ``\dataset{seismic}'' dataset deals with vehicle classification from \dataset{seismic} data
\citep{Duarte_Hu2004}.
The ``\dataset{mushrooms}'' dataset classifies wild mushrooms
into poisonous and non-poisonous varieties based on their physical characteristics \citep{Iba_etal1988}.
The ``\dataset{phishing}'' dataset uses characteristics of a web address to predict whether
or not a website is being used for nefarious purposes \citep{Mohommad_etal2014}.

The final dataset, the ``\dataset{humidity}'' dataset, 
comprises simulated satellite radiometer radiances across 7 different frequencies in the microwave range.
Corresponding to each instance is a value for relative humidity at a single
vertical level.
These humidity values have been discretized into 8 ranges to convert it into a statistical classification problem.
A full description of the genesis of this dataset as well as a rationale for
treatment using statistical classification is contained in \citet{Mills2009}.
The statistical classification methods discussed in this paper were originally
devised specifically for this problem.

Most of the datasets have been supplied already divided into a ``test'' set and a ``training'' set.
If this is the case, then it is noted in the summary in Table \ref{summary}
and the data has been used as given with the training set used for training
and the test set used for testing.
If the data is provided all in one lump, then it was randomly divided into 
test and training sets with the division different for each of the ten numerical trials.

To provide the best idea of when the technique is effective and when it is
not, results from all 17 datasets will be shown. 
All datasets were pre-processed in the same way: by taking the averages and
standard deviations of each feature from the training data and subtracting
the averages from both the test and training data and dividing by the 
standard deviations.
Features that took on all the same value in the training data were removed.

\section{A simple example}

\label{example}

\begin{figure}
  \ifsubmit
    \includegraphics[width=0.9\textwidth]{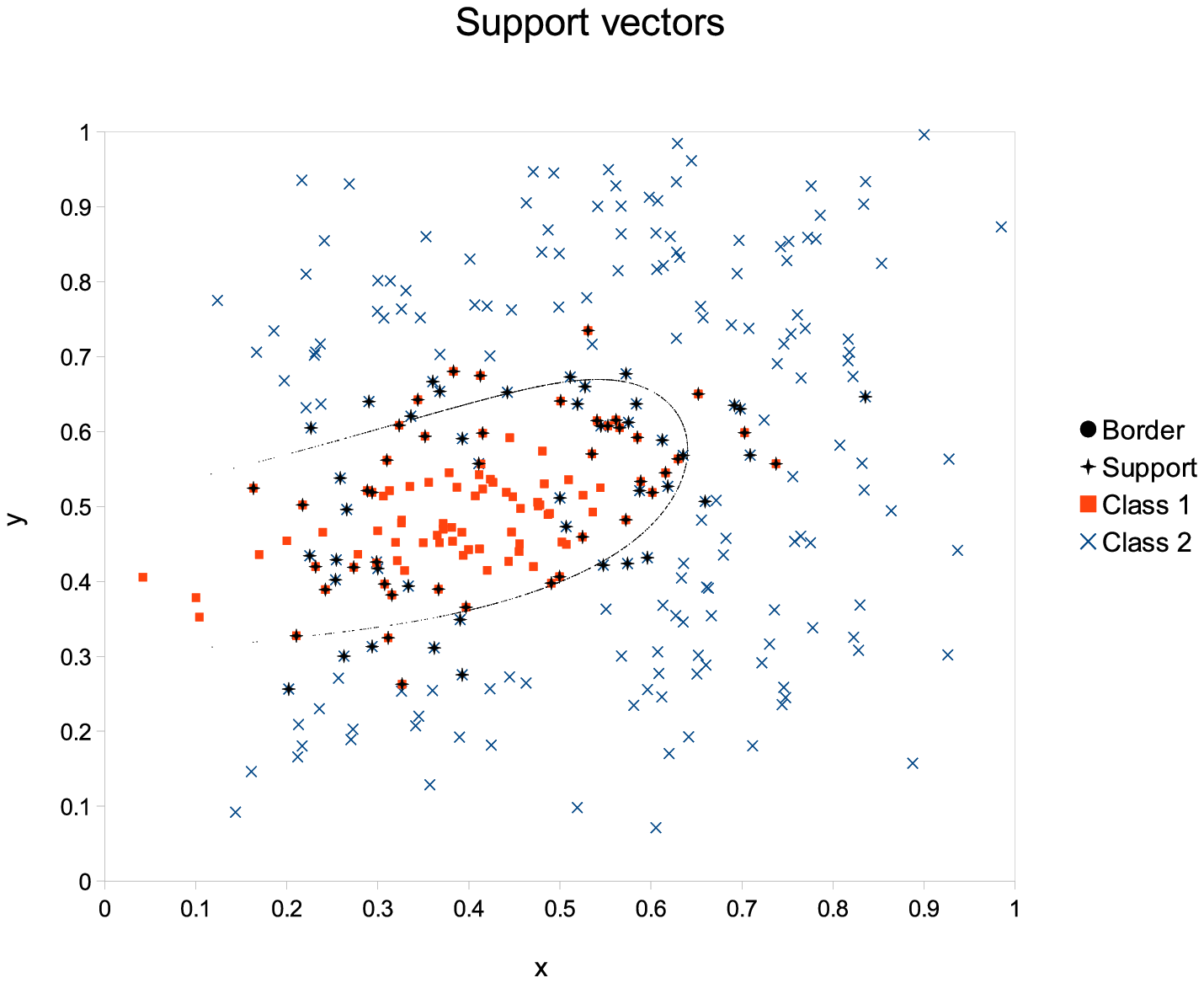}
  \else
    \includegraphics[width=0.9\textwidth]{../support_vectors}
  \fi
  \caption{Support vectors for a pair of synthetic test classes.}\label{sample_sv}
\end{figure}

We use the pair of synthetic test classes defined in \citet{Mills2011} to 
illustrate the difference between support vectors and border vectors.
Figure \ref{sample_sv} shows a realization of the two sample classes 
in red and blue, comprising 300 samples total, along
with the support vectors derived from a LIBSVM model.
The support vectors are a subset of the training samples and while they
tend to cluster around the border, they do not define it.
For reference, the border between the two classes is also shown.
This has been derived from the border-classification method described in 
Section \ref{border_method} using the mathematical definition of the classes,
hence it represents the ``true'' border to within a very small numerical error.

\begin{figure}
\includegraphics[width=0.9\textwidth]{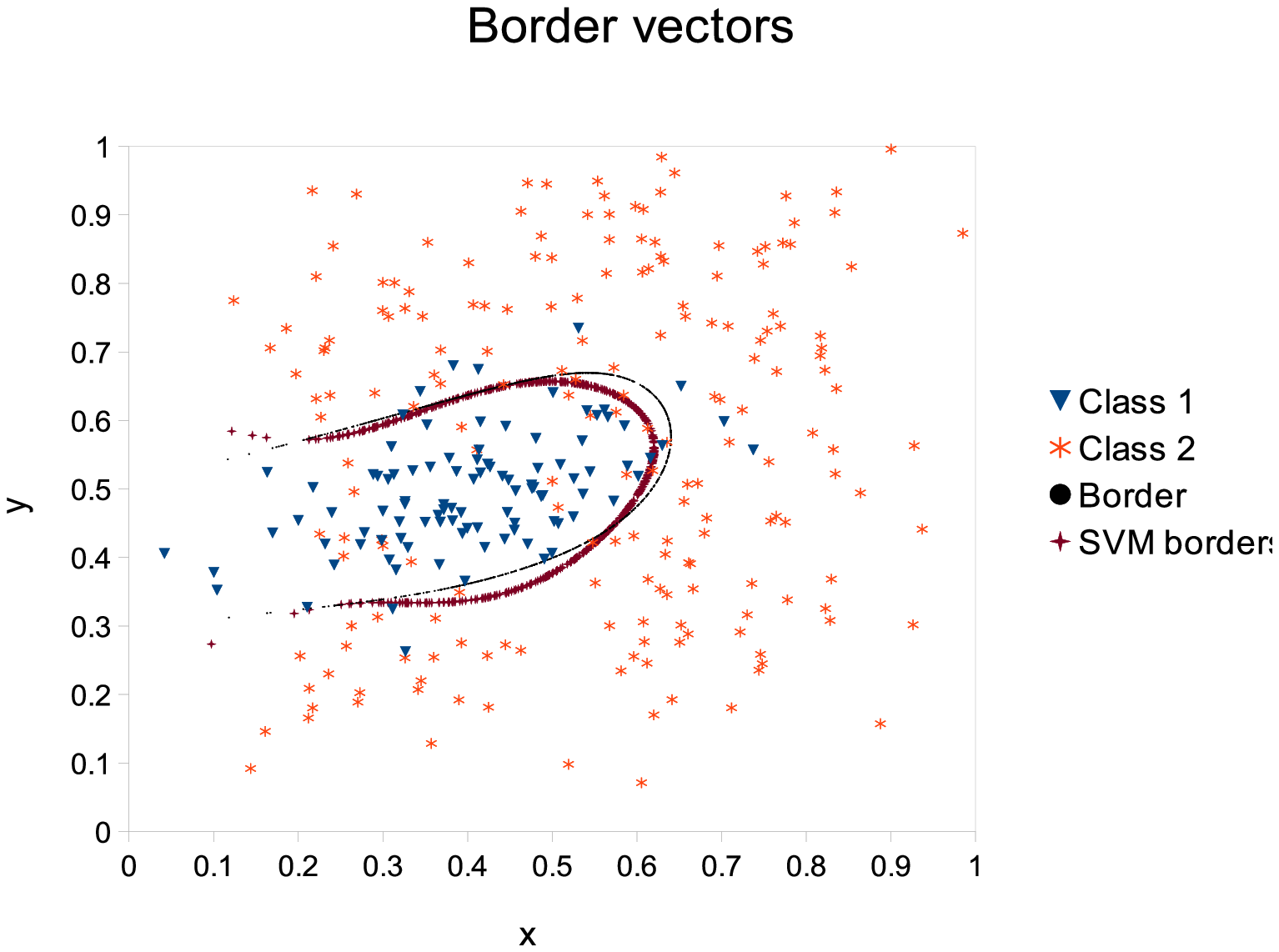}
\caption{Borders mapped by the border-classification method starting with probabilities from the class definitions and a support vector machine (SVM).}
\label{border_vectors}
\end{figure}

\begin{figure}
\includegraphics[width=0.9\textwidth]{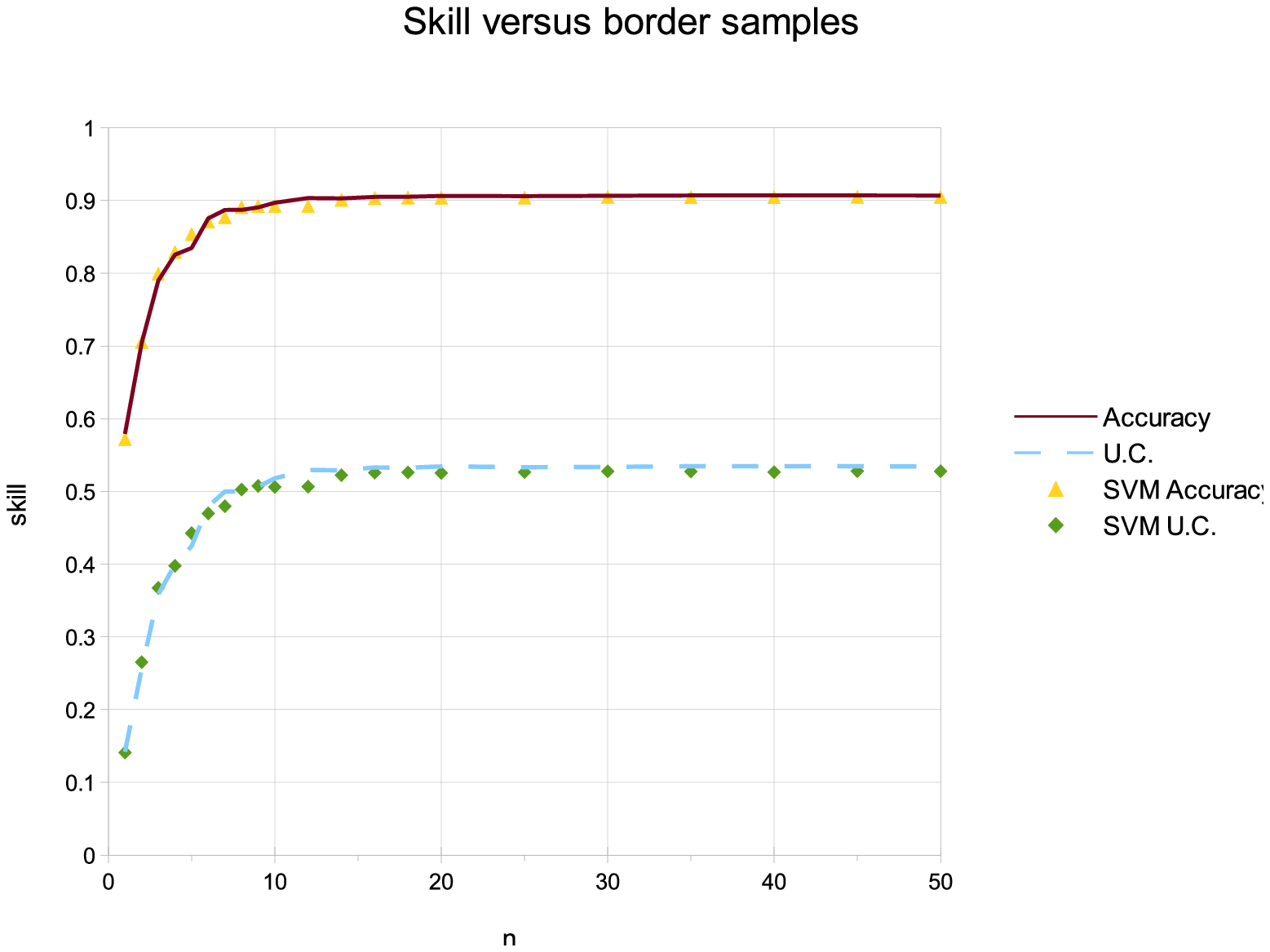}
\caption{Classification accuracy and uncertainty coefficient for border-classification starting with probabilities from the class definitions, and a support vector machine (SVM). Average of 20 trials.}
\label{skill_v_nb}
\end{figure}

The true border is also compared with those derived from LIBSVM
probability estimates in Figure \ref{border_vectors}.
The classes are again shown for reference.
While these borders contain several hundred samples for a clear view of where
they are located using each method, in fact the method works well with
surprisingly few samples.  Figure \ref{skill_v_nb} shows a plot of the skill
versus the number of border samples, where {\it U.C.} stands for
uncertainty coefficient. Note that the scores saturate at only about 20
samples meaning that for this problem at least, very fast classifications are
possible.

Unlike support vectors, the number of border samples required is approximately
independent of the number of training samples.
In addition to skill as a function of border samples,
Figure \ref{skill_v_nb} shows skill as a function of the number of border 
samples for a both an SVM-trained border classifier as well as
a border classifier trained from the mathematical definition of the 
classes themselves.
Note that both curves are roughly the same.
So long as the complexity of
the problem does not increase, adding new training samples does not increase
the number of border samples required for maximum accuracy.

\begin{figure}
  \ifsubmit
    \includegraphics[width=0.9\textwidth]{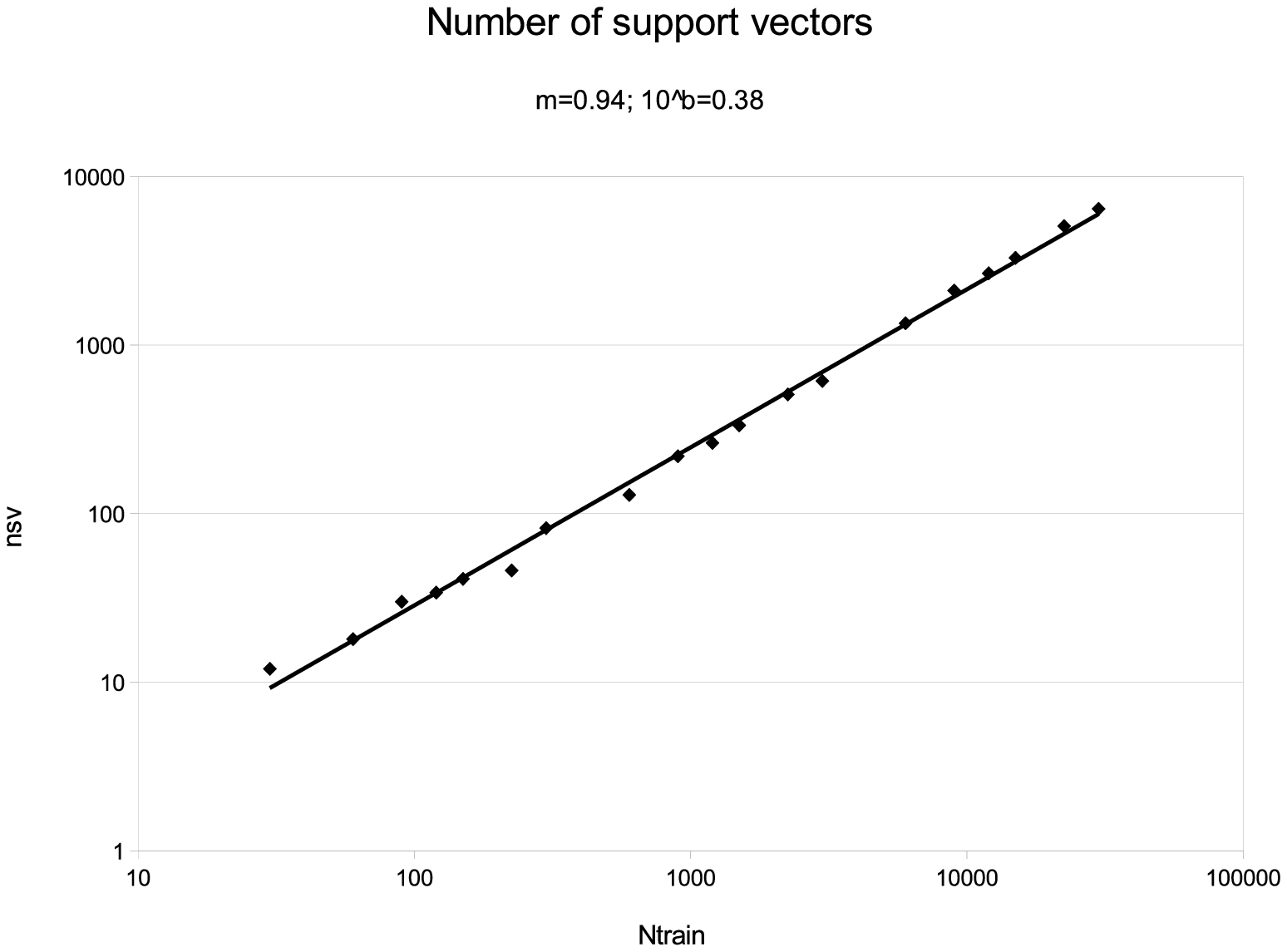}
  \else
    \includegraphics[width=0.9\textwidth]{../nsv}
  \fi
\caption{Number of support vectors against number of training vectors for pair of synthetic test classes. Fitted curve returns exponent of 0.94 and multiplication coefficient of 0.38.}
\label{nsv}
\end{figure}

Figure \ref{nsv} shows the number of support vectors versus the
number of training samples. The fitted curve is approximately linear 
with an exponent of 0.94 and multiplication coefficient of 0.38.
In other words, for this problem there will be approximately 38 \% as many 
support vectors as there are training vectors.

\begin{figure}
  \ifsubmit
    \includegraphics[width=0.9\textwidth]{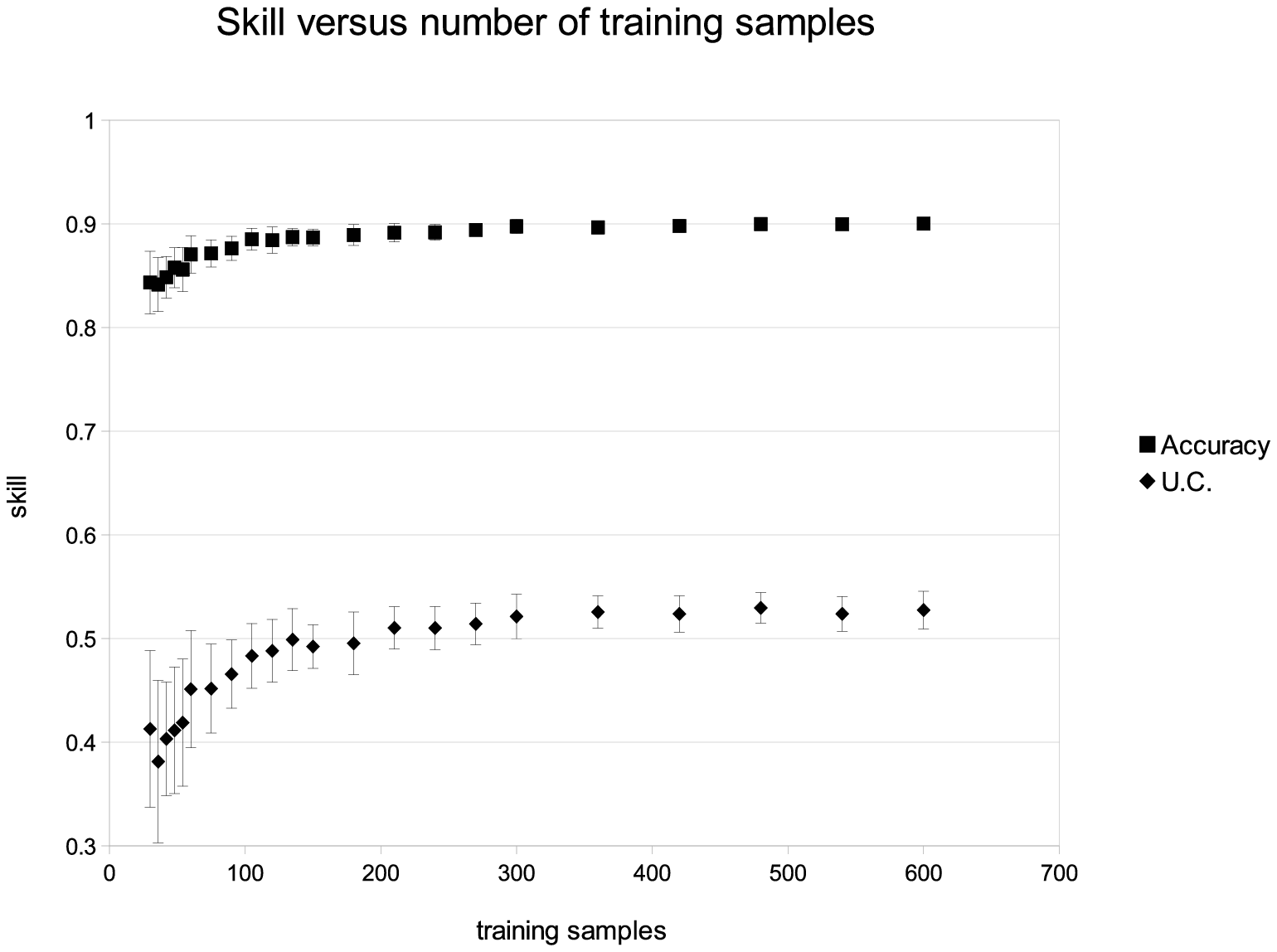}
  \else
    \includegraphics[width=0.9\textwidth]{../skill_v_nt}
  \fi
\caption{Classification accuracy and uncertainty coefficient for a support vector machine (SVM) trained with different numbers of samples.
Error bars represent the standard deviation of 20 trials.}
\label{skill_v_nt}
\end{figure}

Of course it's possible to speed up an SVM by sub-sampling the training data
or the resulting support vectors.
In such case, the sampling must be done carefully so as not to reduce the
accuracy of the result.
Figure \ref{skill_v_nt} shows the effect on classification skill for the
synthetic test classes when the number of training samples is reduced.
Skill scores start to saturate at between 200 and 300 samples.
By contrast, Figure \ref{skill_v_nb} implies that you need only 20 border samples
for good accuracy, so even with only 200 training samples you will still
have improved efficiency by using the borders technique.

This suggests a simple scaling law. The number of training samples required
for good accuracy, and hence the number of support vectors, 
should be proportional to the approximate volume occupied by the
training data in the feature space: $\mintrain \propto \featurevolume$ where 
$\mintrain$ is the minimum number of training vectors and $\featurevolume$ is volume.
Then the number of border vectors should be proportional to the volume
taken to the root of the dimension of the feature space 
then raised to the power of one less:
$\minborders \propto \featurevolume^\frac{1}{\dimension-1}$.
Putting it together, we can relate the two as follows:
\footnote{The correct form of the relation is: $\minborders \propto \mintrain^\frac{\dimension-1}{D}$.
This footnote does not appear in the original article.
}

\begin{eqnnon}
	\minborders \propto \mintrain^\frac{1}{\dimension-1}
	\label{scaling_law}
\end{eqnnon}

where $\minborders$ is the minimum number of border vectors required for good
accuracy.

\begin{figure}
  \ifsubmit
    \includegraphics[width=0.9\textwidth]{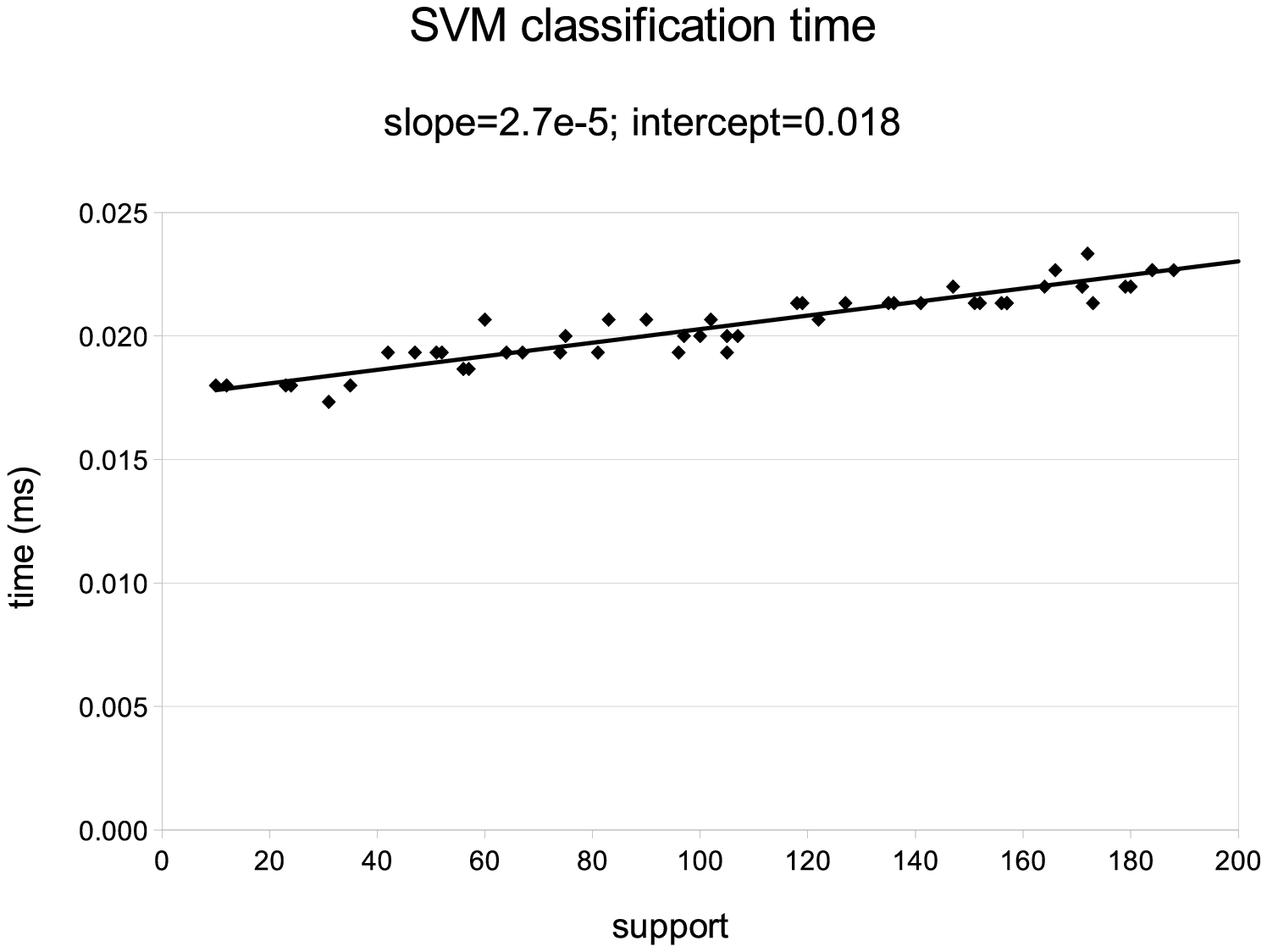}
  \else
    \includegraphics[width=0.9\textwidth]{../svm_time}
  \fi
\caption{Classification time for a SVM for a single test point versus number of support vectors.}
\label{svm_time}
\end{figure}

In other words, provided the class borders are not fractal \citep{Ott1993}, 
mapping only the border between classes should always be
faster than techniques that map the entirety of the class locations,
{\correction although the advantage becomes less the larger the dimension of the 
feature space.}
This includes kernel density methods including SVM as well
as similar methods such as learning vector quantization (LVQ) 
\citep{Kohonen2000,LVQ_PAK}
that attempt to create an idealized representation of the classes through
a set of ``codebook'' vectors.

\begin{figure}
  \ifsubmit
    \includegraphics[width=0.9\textwidth]{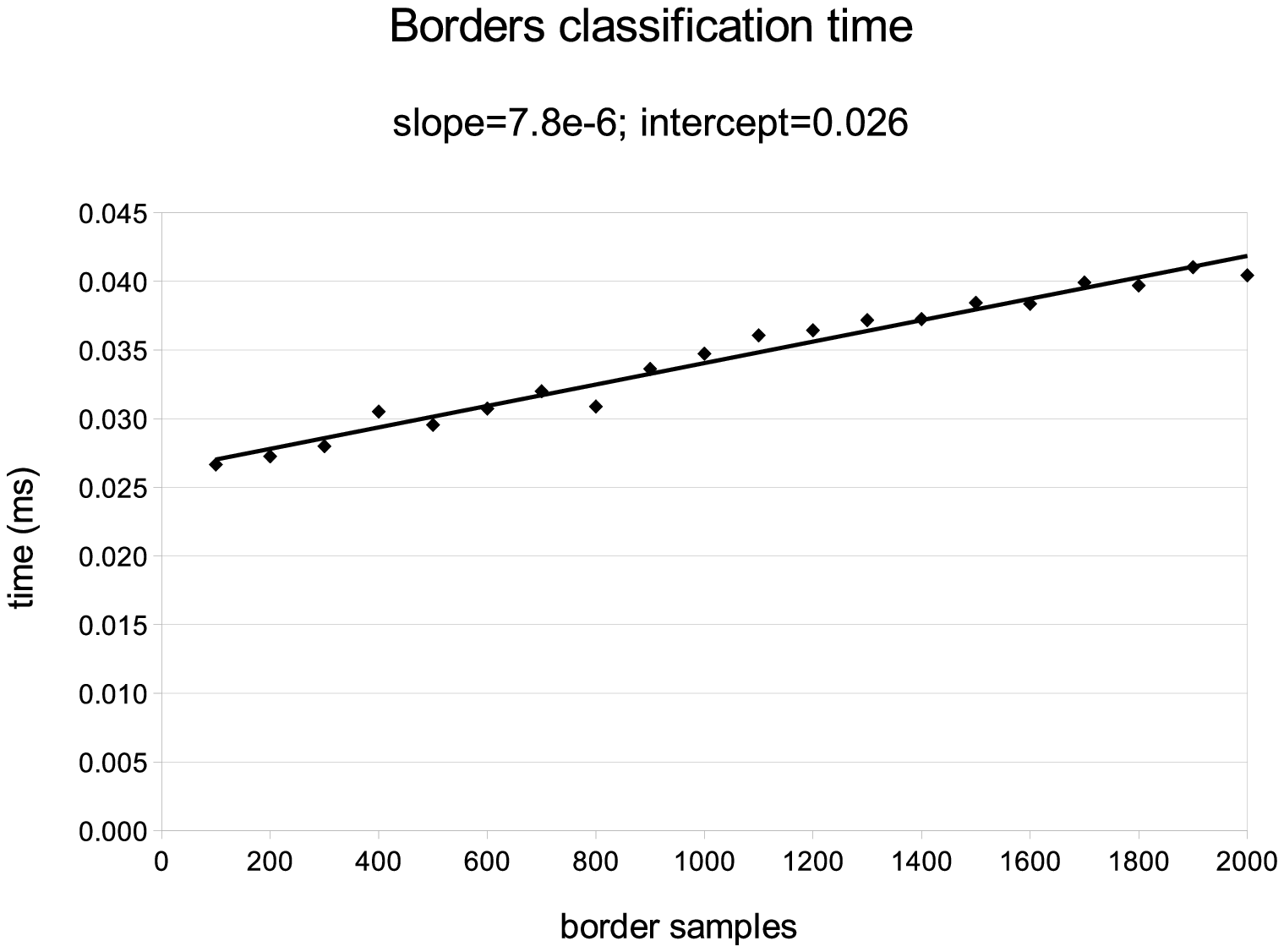}
  \else
    \includegraphics[width=0.9\textwidth]{../border_time}
  \fi
\caption{Classification time for a border classifier for a single test point versus number of border samples.}
\label{border_time}
\end{figure}

To make this more concrete, Figure \ref{svm_time}
plots the classification time versus the number of support vectors
for a SVM
while Figure \ref{border_time} plots the classification time
versus the number of border samples for a border classifier.
Classification times are for a single test point.
Fitted straight lines are overlaid for each and the slope and intercept 
printed in the subtitle.

\begin{figure}
  \ifsubmit
    \includegraphics[width=0.9\textwidth]{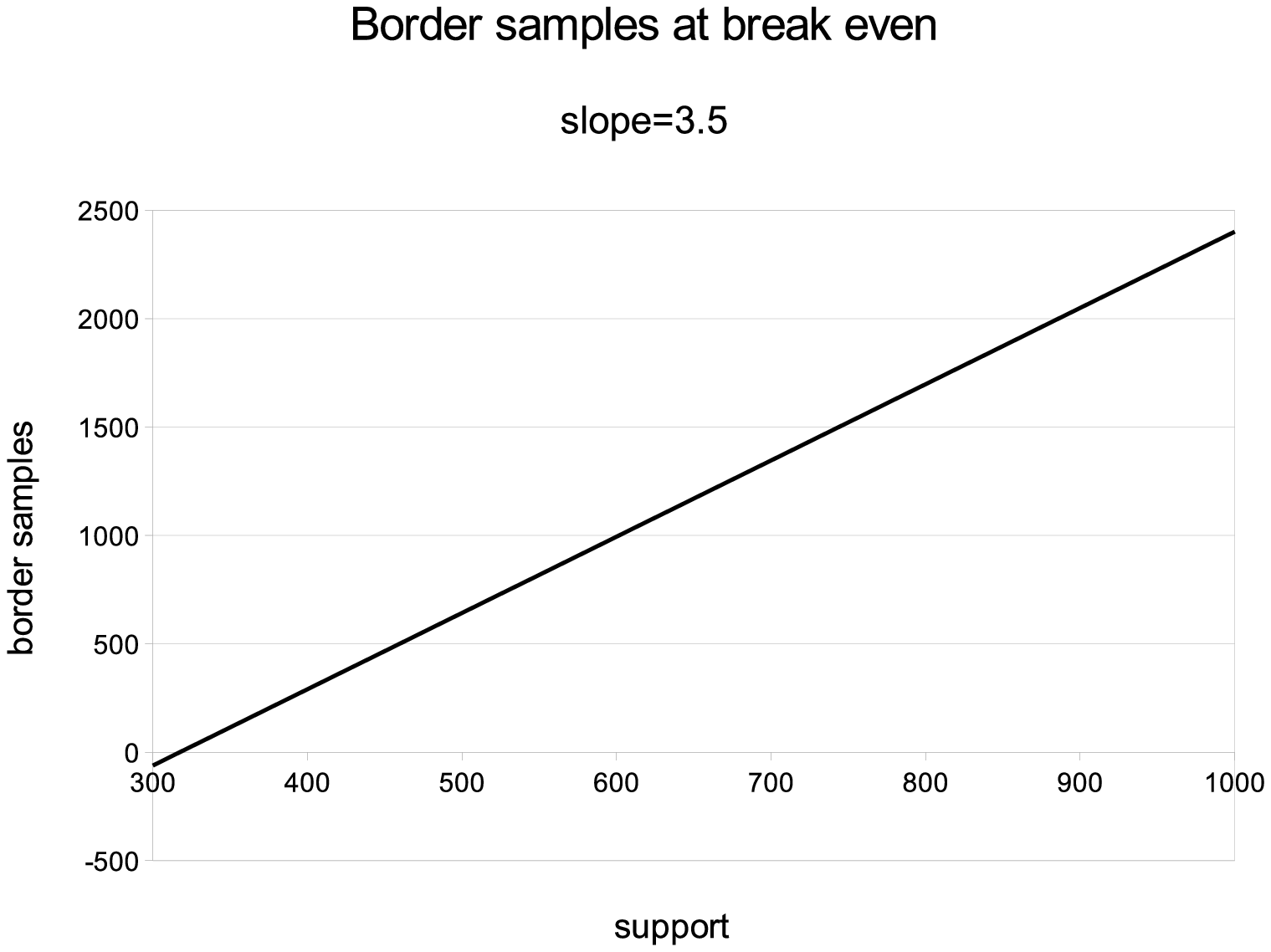}
  \else
    \includegraphics[width=0.9\textwidth]{../break_even}
  \fi
\caption{Number of border samples versus number of support vectors for equal classification times.}
\label{break_even}
\end{figure}

Figure \ref{break_even} plots the number of border vectors versus the number
of support vectors at the ``break even'' point: that is, the classification
time is the same for each method.
This graph was simply derived from the fitted coefficients of the previous
two graphs.
It is somewhat optimistic
since LIBSVM has a larger overhead than the border classifiers.
This overhead would be less significant for larger problems 
with the ``rule of thumb'' suggested by the slope 
that the number of border vectors should be less than three times the support
for a reasonable gain in efficiency.

Unfortunately the graph is not general: while the borders method scales linearly with
the number of classes, in LIBSVM there is some shared calculation for multi-class problems.
That is, some of the support vectors are shared between classes moreover the number will be different for each problem.
Model size comparisons between the two methods should ideally be between the total 
number of support vectors versus the total number of border vectors, not border (or support) vectors per class.
Both methods will tend to scale linearly with the number of attributes, with a small
component independent but a different amount for each method.
Once we take into account the number of classes and number of attributes, the
model for time complexity becomes quite complex so no attempt will be made here
to fit it.

\section{Case studies}

\label{results_section}

\begin{table}
	\caption{Summary of the parameters used in the numerical trials for each of the three methods: Linear, SVM (support vector machine) and ACC (``accelerated'' SVM).}
	\label{param}
        
\begin{tabular}{|l|ll|l|ll|l|}
	\hline
	Dataset	& Stat. & & Linear & SVM & & Accel. \\\hline
	& trials & $\datafraction$ & type$^+$ & $\gamma$ & C & $\nborder$ \\\hline\hline
	heart & 10 & 0.4 & 1 & 0.01 & 0.5 & 100 \\
	shuttle & $10^*$ & 0.25 & 4 & 0.111 & 1 & 100 \\
	sat & $10^*$ & 0.31 & 4 & 0.1 & 50 & 200 \\
	segment & 10 & 0.4 & 4 & 0.1 & 100 & 50 \\
	dna & $10^*$ & 0.372 & 2 & 0.0055 & 1 & 1000 \\
	splice & $10^*$ & 0.685 & 1 & 0.00167 & 1 & 500 \\
	codrna & $10^*$ & 0.82 & 4 & 0.125 & 1 & 500 \\
	letter & $10^*$ & 0.4 & 4 & 0.065 & 1 & 75 \\
	pendigits & $10^*$ & 0.318 & 4 & 0.01 & 50 & 200 \\
	usps & $10^*$ & 0.216 & 4 & 0.004 & 1 & 50 \\
	mnist & 1 & 0.143 & 1 & 0.0015 & 50 & 500 \\
	ijcnn1 & $10^*$ & 0.647 & 0 & 0.045 & 1 & 500 \\
	madelon & $10^*$ & 0.231 & 2 & 0.002 & 1 & 100 \\
	seismic & $10^*$ & 0.2 & 1 & 0.02 & 1 & 200 \\
	mushrooms & 10 & 0.4 & 1 & 0.0089 & 50 & 200 \\
	phishing & 10 & 0.4 & 0 & 0.00147 & 1 & 500 \\
	humidity & $10^*$ & 0.4 & 5 & 0.143 & 50 & 200 \\
	\hline
\end{tabular}

	\vspace{1 ex}

	\raggedright 
	$^*$ Some operations received only a single trial. See text.

	$^+$ Key:\citep{Fan_etal2008}
	\begin{enumerate}
			\setcounter{enumi}{-1}
		\item L2-regularized logistic regression, primal optimization
		\item L2-regularized L2-loss SVM, dual optimization
		\item L2-regularized L2-loss SVM, primal optimization
		\item L2-regularized L1-loss SVM, dual optimization
		\item SVM by Crammer and Singer \citep{Crammer_Singer2002}
		\item L1-regularized L2-loss SVM
	\end{enumerate}
\end{table}

\begin{table}
	\caption{Collation of results for numerical trials of the three different statistical classification methods over seventeen different datasets.}
	\label{results}
	{\small
		\begin{tabular}{|ll|lll|}
\hline
dataset & quantity & Linear & SVM & Accel. \\
\hline\hline
heart & train (s) & $\mathbf{       0.023\pm    0.013}$ & $       0.062\pm    0.015$ & $        0.060\pm   0.009$\\
 & test (s)       & $\mathbf{       0.021\pm    0.011}$ & $        0.030\pm   0.009$ & $       0.027\pm    0.012$\\
 & acc       & $       0.827\pm    0.024$ & $\mathbf{       0.829\pm    0.022}$ & $       0.829\pm    0.022$\\
 & U.C.      & $0.337\pm    0.054$ & $\mathbf{0.341\pm    0.054}$ & $       0.341\pm    0.054$\\
\hline
shuttle & train (s) & $        96\pm      7$ & $\mathbf{        88\pm      4}$ & $        2.33\pm    0.05$\\
	       & test (s)       & $\mathbf{0.22\pm    0.01}$ & $        5.1\pm     0.1$ & $2.60\pm    0.05$\\
 & acc       & $       0.813$ & $\mathbf{       0.998}$ & $       0.996\pm   0.001$\\
 & U.C.      & $0.607$ & $\mathbf{0.981}$ & $       0.979\pm   0.005$\\
\hline
sat & train (s) & $\mathbf{       0.99\pm    0.05}$ & $        8.7\pm     0.4$ & $        8.1\pm     0.2$\\
    & test (s)       & $       \mathbf{0.083\pm    0.017}$ & $        1.01\pm    0.07$ & $0.85\pm     0.03$\\
 & acc       & $       0.827$ & $\mathbf{       0.914}$ & $       0.889\pm   0.004$\\
 & U.C.      & $0.679$ & $\mathbf{0.800}$ & $       0.764\pm   0.007$\\
\hline
segment & train (s) & $\mathbf{       0.12\pm    0.03}$ & $       0.78\pm    0.05$ & $       0.98\pm    0.04$\\
 & test (s)       & $\mathbf{       0.034\pm    0.018}$ & $       0.25\pm    0.02$ & $       0.235\pm    0.015$\\
 & acc       & $       0.933\pm   0.008$ & $\mathbf{       0.962\pm   0.009}$ & $       0.956\pm   0.006$\\
 & U.C.      & $0.881\pm    0.014$ & $\mathbf{0.917\pm    0.017}$ & $       0.909\pm    0.013$\\
\hline
dna & train (s) & $\mathbf{       0.52\pm    0.03}$ & $        18.2\pm      1.1$ & $        21.3\pm     0.4$\\
 & test (s)       & $\mathbf{       0.13\pm     0.02}$ & $        1.8\pm     0.2$ & $        1.19\pm    0.02$\\
 & acc       & $         0.900$ & $\mathbf{        0.950}$ & $       0.737\pm   0.004$\\
 & U.C.      & $0.684$ & $\mathbf{0.780}$ & $       0.472\pm   0.008$\\
\hline
splice & train (s) & $\mathbf{       0.24\pm    0.04}$ & $        1.9\pm     0.2$ & $        1.6\pm      0.1$\\
       & test (s)       & $\mathbf{0.101\pm   0.009}$ & $       0.64\pm    0.07$ & $0.15\pm    0.02$\\
 & acc       & $       0.846$ & $\mathbf{       0.894}$ & $       0.768\pm   0.003$\\
 & U.C.      & $0.385$ & $\mathbf{0.514}$ & $       0.321\pm   0.008$\\
\hline
codrna & train (s) & $\mathbf{        3.17\pm0.12}$ & $         929$ & $        9.6\pm     0.2$\\
 & test (s)       & $        2.88\pm0.07$ & $         346$ & $\mathbf{         2.70\pm    0.06}$\\
 & acc       & $        0.930$ & $\mathbf{       0.964}$ & $       0.961\pm  0.0004$\\
 & U.C.      & $0.676$ & $\mathbf{0.759}$ & $       0.744\pm   0.002$\\
\hline
letter & train (s) & $\mathbf{        3.00\pm0.08}$ & $        44.8\pm     0.8$ & $        31.9\pm     0.7$\\
 & test (s)       & $\mathbf{       0.22\pm    0.02}$ & $        16.1\pm     0.8$ & $        19.5\pm      1.1$\\
 & acc       & $       0.723$ & $\mathbf{       0.963}$ & $       0.929\pm   0.0022$\\
 & U.C.      & $0.637$ & $\mathbf{0.935}$ & $       0.882\pm   0.003$\\
\hline
pendigits & train (s) & $\mathbf{       0.50\pm    0.03}$ & $        3.71\pm     0.13$ & $        1.85\pm    0.06$\\
 & test (s)       & $\mathbf{         0.1\pm    0.012}$ & $        1.24\pm    0.04$ & $        1.12\pm    0.05$\\
 & acc       & $       0.908$ & $\mathbf{       0.978}$ & $       0.975\pm   0.001$\\
 & U.C.      & $0.831$ & $\mathbf{0.950}$ & $       0.945\pm   0.002$\\
\hline
\end{tabular}

	}
\end{table}

\begin{table}
	\caption{Collation of results for numerical trials of the four different statistical classification methods over seventeen different datasets.}
	\label{results2}
	{\small
		\begin{tabular}{|ll|lll|}
\hline
dataset & quantity & Linear & SVM & Accel. \\
\hline\hline
usps & train (s) & $\mathbf{        18.3\pm     0.4}$ & $         104\pm       14$ & $        36.4\pm      1.6$\\
     & test (s)       & $\mathbf{0.305\pm    0.015}$ & $        7.0\pm      1.3$ & $4.1\pm     0.3$\\
 & acc       & $       0.898$ & $\mathbf{       0.949}$ & $        0.940\pm   0.0024$\\
 & U.C.      & $0.783$ & $\mathbf{0.880}$ & $       0.866\pm   0.004$\\
\hline
mnist & train (s) & $     11035$ & $\mathbf{    6381}$ & $    1906\pm  676$\\
 & test (s)       & $\mathbf{        3.64}$ & $         343$ & $         308\pm       24$\\
 & acc       & $ 0.915$ & $\mathbf{       0.972}$ & $        0.960\pm  0.0008$\\
 & U.C.      & $ 0.806$ & $       \mathbf{0.924}$ & $         0.900\pm   0.002$\\
\hline
ijcnn1 & train (s) & $\mathbf{        1.00\pm0.04}$ & $         560$ & $        7.52\pm     0.18$\\
 & test (s)       & $\mathbf{        1.62\pm0.03}$ & $         109$ & $        2.07\pm    0.04$\\
 & acc       & $       0.632$ & $\mathbf{       0.988}$ & $        0.980\pm   0.001$\\
 & U.C.      & $0.199$ & $\mathbf{0.824}$ & $       0.757\pm   0.008$\\
\hline
madelon & train (s) & $\mathbf{        1.04\pm    0.052}$ & $        86\pm      6$ & $        17.7\pm     0.4$\\
 & test (s)       & $       0.193\pm    0.016$ & $        5.11\pm     0.14$ & $\mathbf{       0.094\pm    0.015}$\\
 & acc       & $       0.578$ & $       0.578$ & $\mathbf{        0.580\pm   0.007}$\\
 & U.C.      & $0.0178$ & $      0.0178$ & $\mathbf{0.0243\pm   0.004}$\\
\hline
seismic & train (s) & $\mathbf{        4.8\pm0.2}$ & $    55905$ & $          74.0\pm     0.6$\\
	       & test (s)       & $\mathbf{0.70\pm0.03}$ & $         350$ & $2.37\pm    0.05$\\
 & acc       & $       0.648$ & $\mathbf{       0.724}$ & $       0.695\pm   0.004$\\
 & U.C.      & $0.292$ & $\mathbf{0.308}$ & $       0.275\pm   0.004$\\
\hline
mushrooms & train (s) & $\mathbf{       0.454\pm    0.016}$ & $        33.1\pm     0.8$ & $        2.81\pm    0.07$\\
 & test (s)       & $       0.229\pm    0.014$ & $        3.5\pm      0.1$ & $\mathbf{       0.18\pm    0.02}$\\
 & acc       & $\mathbf{           1.\pm  0.0003}$ & $           1.\pm  0.0003$ & $       0.998\pm   0.002$\\
 & U.C.      & $\mathbf{       0.999\pm   0.003}$ & $       0.996\pm    0.003$ & $       0.985\pm    0.017$\\
\hline
phishing & train (s) & $\mathbf{       0.52\pm    0.03}$ & $        39.3\pm     0.9$ & $        4.34\pm    0.09$\\
	& test (s)       & $       \mathbf{0.22\pm    0.02}$ & $        4.1\pm     0.1$ & $0.32\pm    0.02$\\
 & acc       & $       0.939\pm   0.002$ & $\mathbf{       0.958\pm   0.003}$ & $       0.952\pm   0.003$\\
 & U.C.      & $0.664\pm   0.009$ & $\mathbf{0.747\pm    0.012}$ & $       0.721\pm    0.012$\\
\hline
humidity & train (s) & $\mathbf{        16.4\pm1.8}$ & $    2712$ & $        71.1\pm      1.6$\\
	& test (s)       & $\mathbf{0.55\pm0.03}$ & $         236$ & $9.2\pm      0.2$\\
 & acc       & $       0.441$ & $\mathbf{       0.609}$ & $       0.605\pm  0.0009$\\
 & U.C.      & $0.292$ & $\mathbf{0.479}$ & $       0.474\pm  0.0005$\\
\hline
\end{tabular}

	}
\end{table}

Three classification models were tested on each of the 17 datasets described in
Section \ref{datasets}: a basic linear classifier, 
a support vector machine (SVM) and a borders
model derived from the previous SVM model (Accel. for ``accelerated'' SVM).
SVM should be more accurate than linear classification for problems in which
the classes are not linearly separable.
Meanwhile, for large problems, the borders technique should produce significant time savings over SVM while having little effect on accuracy.

The parameters used for each method are summarized in Table \ref{param}.
The parameter, $\nborder$, in the borders technique
was chosen for the best compromise between reduced accuracy and a speed
improvement over SVM. 

For SVM, a Gaussian kernel, also known as a radial basis function (RBF), is used:
\begin{eqnnon}
	K(\vec x, \vec y) = \exp \left ( - \rbkernelparam | \vec y - \vec x |^2 \right )
\end{eqnnon}
where $\rbkernelparam$ is a tunable parameter. 
$\svmcost$ is a cost parameter added to reduce over-fitting: see
Equations (\ref{dual_problem}) and (\ref{constraint1}).

In order to get a confidence interval on the results,
ten trials were performed for most of the datasets.
In some cases, only a single trial was performed either because the operation
took too long or because of a pre-existing separation between test and training
data which was taken ``as is''. Single trials are indicated through the absence of
error bars which are calculated from the standard deviation.
$\datafraction$ is the fraction of test data relative to the total number
of samples.

The results are summarized in Tables \ref{results} and \ref{results2} including training
and test time for each method as well as skill scores. There are two skill scores,
the first being simple accuracy or fraction of correct guesses while the second,
called the uncertainty coefficient, is based on information entropy and is described in Section \ref{skill_scores}.
The best values for each dataset are highlighted in bold.
The SVM-borders method is not a stand-alone method thus its training time is never
highlighted.

\label{discussion}

\begin{table}
	\caption{Total number of support vectors versus total number of border samples.}
	\begin{tabular}{|l|lllllll|}
	\hline
	& $\dimension$ & $\nclass$ & $\nsample$ & Total & Total & Time (s) & Time (s)\\
	& & & & support & borders & SVM & accel. \\\hline
	heart & 13 & 2 & 162 & $111\pm5$ & 100 & 0.030 & 0.027\\
	shuttle & 9 & 7 & 43500 & 1277 & 2100 & 5.1 & 2.60\\
	sat & 36 & 6 & 4435 & 1447 & 3000 & 1.01 & 0.85 \\
	segment & 19 & 7 & 1386 & $340\pm11$ & 1050 & 0.25 & 0.24 \\
	dna & 180 & 3 & 2000 & 1288 & 1000 & 1.8 & 1.19 \\
	splice & 60 & 2 & 1000 & 678 & 500 & 0.64 & 0.15 \\
	codrna & 8 & 2 & 59535 & 8978 & 500 & 346 & 2.70 \\
	letter & 112 & 26 & 12000 & 5355 & 5850 & 16.1 & 19.5 \\ 
	pendigits & 16 & 10 & 7494 & 602 & 2250 & 1.24 & 1.12 \\
	usps & 256 & 10 & 7291 & 2081 & 2250 & 7.0 & 4.11 \\
	mnist & 665 & 10 & 45000 & 12282 & 22500 & 343 & 308 \\
	ijcnn1 & 22 & 2 & 49990 & 4888 & 500 & 109 & 2.07 \\
	madelon & 500 & 2 & 2000 & 1959 & 100 & 5.11 & 0.094 \\
	seismic & 50 & 2 & 78823 & 44385 & 200 & 350 & 2.37 \\
	mushrooms & 112 & 2 & 4874 & $1054\pm14$ & 200 & 3.5 & 0.18 \\
	phishing & 68 & 2 & 6633 & $1445\pm20$ & 500 & 4.1 & 0.32 \\
	humidity & 7 & 8 & 51840 & 37796 & 5600 & 236 & 9.2 \\ 
	\hline
\end{tabular}

	\label{sum_nsv}
\end{table}

\begin{table}
  \caption{Results from SVM trials after sub-sampling to match the SVM-borders
	  uncertainty coefficient score is matched.}
  \label{subsampling_table}
  {\small
    \begin{tabular}{|l|llllll|}
\hline
dataset & samples & S.V. & train (s) & test (s) & accuracy & U.C. \\
\hline
shuttle  & $    6522$ & $         518\pm      8$ & $        3.6\pm      0.2$ & $        3.60\pm    0.07$ & $       0.997\pm  0.001$ & $       0.972\pm   0.008$\\
sat  & $    1772$ & $         733\pm       13$ & $        2.24\pm    0.08$ & $       0.70\pm    0.04$ & $       0.893\pm   0.007$ & $       0.766\pm     0.010$\\
segment  & $    1037\pm     1$ & $         290\pm        9$ & $       0.53\pm    0.03$ & $       0.229\pm   0.009$ & $       0.954\pm   0.007$ & $       0.903\pm    0.012$\\
dna  & $          99$ & $        98.9\pm     0.3$ & $       0.31\pm    0.03$ & $       0.30\pm    0.01$ & $       0.835\pm    0.012$ & $       0.478\pm    0.032$\\
splice  & $         169$ & $         158\pm      3$ & $         0.20\pm    0.02$ & $       0.29\pm     0.03$ & $       0.826\pm    0.012$ & $       0.334\pm    0.026$\\
codrna  & $    14883$ & $    2781\pm       51$ & $        51\pm      4$ & $         114\pm      3$ & $       0.961\pm  0.0006$ & $       0.741\pm   0.003$\\
pendigits  & $    4864$ & $         515\pm       12$ & $        2.23\pm    0.06$ & $        1.21\pm    0.05$ & $       0.975\pm   0.002$ & $       0.944\pm   0.003$\\
usps  & $    3276$ & $    1255\pm       21$ & $        31\pm      2$ & $        4.6\pm     0.3$ & $       0.941\pm   0.003$ & $       0.865\pm   0.006$\\
mnist & $ 22498$ & $7729$ & $2746$ & $253$ & $0.965$ & $0.909$ \\
ijcnn1  & $    12596\pm  315$ & $    1805\pm       52$ & $          40\pm      3$ & $          34\pm      1$ & $       0.983\pm   0.001$ & $       0.744\pm    0.012$\\
seismic  & $    3934$ & $     2499\pm       99$ & $        29\pm      3$ & $        16\pm      4$ & $       0.705\pm   0.003$ & $       0.275\pm   0.005$\\
mushrooms  & $    1218$ & $         398\pm       17$ & $         2.4\pm     0.3$ & $       0.94\pm    0.07$ & $       0.997\pm   0.001$ & $       0.976\pm    0.012$\\
phishing  & $    2652.2\pm     0.4$ & $         769\pm       11$ & $        7.2\pm     0.8$ & $        1.5\pm     0.1$ & $       0.952\pm   0.003$ & $       0.719\pm    0.012$\\
humidity  & $    20637\pm     1$ & $    15162\pm       73$ & $         331\pm        9$ & $        83\pm      3$ & $       0.603\pm   0.002$ & $       0.471\pm    0.002$\\
\hline
\end{tabular}

  }
\end{table}

All but one of the classification problems show a significant speed increase with the application
of the borders technique with \dataset{letter} being the lone exception.
Despite its simplicity, linear classification is shown to be competitive with
a non-parametric method such as SVM for many of the problems and it is almost 
always faster both for training and for testing.
(For the author, at least, this is one of the more surprising results.)

Table \ref{sum_nsv} is an attempt to get a handle on the relative time complexity 
of the two methods--SVM and SVM-borders--and lists all the relevant variables: number of features,
number of classes, total number of training samples, total support for SVM,
total number of border samples for the borders method compared with the resulting
classification time for the two methods. The two most relevant variables here are the
number of support vectors versus the number of border vectors.
In order to get a successful speed increase, the former should be larger than 
the latter, but as is apparent from some problems such as \dataset{shuttle}, 
\dataset{pendigits}, and \dataset{usps},
even having more border samples can sometimes produce a significant,
although modest, improvement.

All increases in speed, however, come at the cost of accuracy.
The question is, is the speed increase worth the decrease in skill?
To test this, we sub-sample the datasets and then re-apply the SVM training
until the skill the two methods, SVM and SVM-borders, as measured by
the uncertainty coefficient matches.

It might seem more expedient to directly sub-sample the support vectors themselves
rather than the training data.
This, however, was found not to work and generated a precipitous drop in accuracy. 
Since the sparse coefficient set, $\vec \svmcoeff$, is found through simultaneous 
optimization, the support vectors turn out to be interdependent.

Depending on how much the dataset is reduced, sub-sampling should be done with at
least some care. 
On one hand, a more sophisticated sub-sampling technique might be considered a method on its own, 
comparable with the borders technique, but also likely requiring multiple training phases using the
original technique thus making it significantly slower.
On the other hand, at minimum we should consider the relative size of each class distribution.
If there are roughly the same number of classes, then for small sub-samples the relative
numbers should be kept constant.
The \dataset{shuttle} dataset, however, has very uneven class numbers so it was sub-sampled differently
in order to ensure that the smallest classes retain some membership.
Let $\classsize_i$ be the number of samples of the $i$th class.
Then the sub-sampled numbers are given:
\begin{eqnnon}
	\classsize^\prime_i = \subsample(\classsize_i) \classsize_i
	\label{subsample}
\end{eqnnon}
The form of $\subsample$ used for the \dataset{shuttle} dataset was:
\begin{eqnnon}
	\subsample(n) = \submultcoef n^{-\subexp}
	\label{subfunction}
\end{eqnnon}
where $\submultcoef=\classsize_1^\subexp$, $0 < \subexp < 1$ is determined based on the
desired total fraction and $\classsize_1$ is the number of samples in the 
smallest class.
To understand how this functional form was chosen, please see Appendix \ref{shuttle_subsampling}.

The results of the sub-sampling exercise are shown in Table \ref{subsampling_table}.
This gives us a clearer understanding of whether or not and
when SVM acceration through borders sampling is effective.

There are at least two major sources of error for the borders technique.
First, it provides only limited sampling of the discrimination
border and this sampling is not strongly optimized.
The sampling method, using pairs of training points of opposite class, will
tend to favour regions of high density, however  
directly optimizing for classification skill would be the ideal solution.
Second, the probability estimates extrapolate from only a single point.
These two errors will tend to compound, especially after converting to
multiple classes.

\section{Conclusions}

\label{conclusion}

\begin{table}
	\caption{Summary of results for all 17 datasets including a verdict on the success or failure of borders classification to speed up SVM.}
	\label{verdict}
	\begin{tabular}{|l|ll|ll|l|}
	\hline
dataset & time (s) & & U.C. & & verdict \\
 & SVM & accel. & SVM & Accel. & \\
	\hline
heart & $0.030\pm0.009$ & $0.027\pm0.012$ & $0.34\pm0.05$ & $0.34\pm0.05$ & succeeds \\
shuttle & $3.60\pm0.07$ & $2.60\pm0.05$ & $0.972\pm0.008$ & $0.979\pm0.005$ & succeeds \\
	sat & $0.70\pm0.04$ & $0.85\pm0.03$ & $0.77\pm0.01$ & $0.76\pm0.007$ & fails \\
segment & $0.229\pm0.009$ & $0.235\pm0.015$ & $0.90\pm0.01$ & $0.91\pm0.01$ & fails \\
	dna & $0.30\pm0.01$ & $1.19\pm0.02$ & $0.48\pm0.03$ & $0.472\pm0.008$ & fails$^*$ \\
	splice & $0.29\pm0.03$ & $0.15\pm0.02$ & $0.33\pm0.03$ & $0.321\pm0.008$ & succeeds$^*$ \\
	codrna & $114\pm3$ & $2.70\pm0.06$ & $0.741\pm0.003$ & $0.744\pm0.002$ & succeeds \\
	letter & $16.1\pm0.8$ & $19.5\pm1.1$ & $0.935$ & $0.882\pm0.003$ & fails \\
	pendigits & $1.21\pm0.05$ & $1.12\pm0.05$ & $0.944\pm0.003$ & $0.945\pm0.002$ & succeeds \\
	usps & $4.6\pm0.3$ & $4.1\pm0.3$ & $0.865\pm0.006$ & $0.866\pm0.004$ & succeeds \\
	mnist & 253 & $308\pm24$ & 0.909 & $0.900\pm0.002$ & fails \\
	ijcnn1 & $34\pm1$ & $2.07\pm0.04$ & $0.74\pm0.01$ & $0.757\pm0.008$ & succeeds \\
	madelon & $5.1\pm0.1$ & $0.09\pm0.02$ & $0.0178$ & $0.0243\pm0.004$ & succeeds \\
	seismic & $16\pm4$ & $2.37\pm0.05$ & $0.275\pm0.005$ & $0.275\pm0.004$ & succeeds \\
	mushrooms & $0.94\pm0.07$ & $0.18\pm0.02$ & $0.98\pm0.01$ & $0.99\pm0.02$ & succeeds$^*$ \\
	phishing & $7.2\pm0.8$ & $0.32\pm0.02$ & $0.72\pm0.01$ & $0.72\pm0.01$ & succeeds \\
	humidity & $83\pm3$ & $9.2\pm0.2$ & $0.471\pm0.002$ & $0.474\pm0.0005$ & succeeds \\
	\hline
\end{tabular}

	\vspace{1 ex}

	\raggedright 
	$^*$ Linear classification generates more accurate results.
\end{table}

The primary goal of this work was to improve the classification efficiency of a SVM
using a simple, piecewise linear classifier which we call the borders classifier.
The outcome for each of the 17 datasets is summarized in Table \ref{verdict}.
When trained from the SVM, the method succeeded for twelve of the datasets.
Of these, two return better results using a linear classifier rather than
borders sampling.
Not a perfect score but certainly worthwhile to try for operational retrievals
where time performance is critical, 
for instance classifying large amounts of satellite data in real time.
This is especially so in light of the high performance ratios for some of the problems:
the \dataset{humidity} dataset is sped up by over 20 times,
for instance, with even higher factors for some of the binary datasets.

The same analysis was not repeated for the linear classifier since 
it is a constant time algorithm, so should always come out the winner.
Even so, in three of the trials, \dataset{codrna}, \dataset{madelon} and
\dataset{mushrooms}, borders classifications was actually faster than a linear
classifier. This may have more to do with implementation than anything else,
however. If raw speed is the main criterion, linear classifiers are
the gold standard. 

Linear classification has the potential to be both fast
and accurate over a broad range of datasets if the features are first mapped
to a higher dimensional space, similar to polynomial regression fitting.
This is done implicitly in SVM \citep{Mueller_etal2001}, but closed-form
feature maps that are also efficient to compute
are difficult to derive for many kernels, including RBF. 
Nonetheless, \citet{Vedaldi_Zisserman2012} have succeeded in doing so for
several kernels popular in computer vision applications.
Sadly, comparative results were not available in time.

It's worthwhile to note where the borders classification algorithm 
is most likely to succeed and conversely where it might fail.
One of the most successful trials was for the \dataset{humidity} dataset which produced one of
the largest time improvements combined with relatively little loss of accuracy.
This makes sense since the method was devised specifically for this problem and
the \dataset{humidity} dataset epitomizes the characteristics for which the technique is most
effective.

Since it assumes that the difference in conditional probabilities is a
smooth and continuous function, the borders method 
tends to work poorly with integer or categorical data as well as problems
with sharply defined, non-overlapping classes.
Indeed, two of the problems where it took the biggest hit in accuracy, \dataset{dna} and \dataset{splice}, 
use binary and categorical data respectively.

Also, since there is no redundancy in calculations for multiple classes, 
whereas in SVM there is considerable redundancy, problems with a large number of
classes should also be avoided.
This can be mitigated by using a multi-class classification method
requiring fewer binary classifiers such as one-versus-the-rest with $O(\nclass)$
performance or a decision tree with $O(\log \nclass)$ performance, rather than
one-versus-one with its $O(\nclass^2)$ time complexity.

The most important characteristic for success with the borders classification 
method is a large number of training samples
used to train a SVM for maximum accuracy.
This also implies a large number of support vectors, making the
SVM slow.
Choosing an appropriate number of border samples allows one to trade off
accuracy for speed, with diminishing returns for larger numbers of border samples.
The borders method, unlike SVM, also has a straightforward interpretation:
the location of the samples represent a hyper-surface that divides the two
classes and their gradients are the normals to this surface.
In this regard it is somewhat similar to rule-based classifiers such as
decision trees.

There are many directions for future work.
An obvious refinement would be to distribute the border samples less 
randomly and cluster them where they are most needed.
As it is, the method of choosing by selecting random pairs of opposite
classes, will tend to distribute them in areas of high density.
The current, random method was found to work well enough.
Another potential improvement would be to position the border samples
so as to directly minimize classification error.
This need not be done all at once as in some of the methods mentioned
in the Introduction, but rather point-by-point to keep the training
relatively fast.
A first guess could be found through a kernel method and then each
pointed shifted along the normal.
Piecewise linear statistical classification methods are simple, 
powerful and fast and we think they should receive more attention.

For certain types of datasets, particularly those with continuum features
data, smooth probability functions (typically overlapping classes) and a large number of samples, 
the borders classification algorithm is an 
effective method of improving the classification time of kernel methods.
Because it is not a stand-alone method, but requires probability estimates,
it can acheive a fast training time since it is not solving a global
optimization problem, yet still maintain reasonable accuracy.
While it may not be the first choice for cracking ``hard'' problems, it
is ideal for workaday problems, such as operational retrievals,
for which speed is critical.

\appendix

\ifsubmit
  \section{Sub-sampling}
\label{shuttle_subsampling}

Let $\classsize_i$ be the number of samples of the $i$th class such
that:
\begin{equation*}
\classsize_i \ge \classsize_{i-1}
\end{equation*}
Let $0 \le \subsample(n) \le 1$ be a function used to sub-sample each of the class
distributions in turn:
\begin{equation*}
\classsize_i^\prime = \subsample(\classsize_i) \classsize_i
\end{equation*}
We wish to retain the rank ordering of the class sizes:
\begin{equation*}
\subsample(\classsize_i) \classsize_i 
\ge \subsample(\classsize_{i-1}) \classsize_{i-1} 
\end{equation*}
while ensuring that the smallest classes have some minimum representation:
\begin{equation}
\subsample(\classsize_i) \le \subsample(\classsize_{i-1})
\label{subsample_constraint1}
\end{equation}
Thus:
\begin{equation*}
	\frac{\mathrm d}{\mathrm d n} \left [ n \subsample(n) \right ] = \subsample(n) + n \frac{\mathrm d \subsample}{\mathrm n} \ge 0
\end{equation*}
\begin{equation}
	\frac{\mathrm d \subsample}{\mathrm d n} \ge - \frac{\subsample(n)}{n}
\label{subsample_constraint2}
\end{equation}
The simplest means of ensuring that both (\ref{subsample_constraint1}) and
(\ref{subsample_constraint2}) are fulfilled, is to multiply the right side
of (\ref{subsample_constraint2}) with a constant, $0 \le \subexp \le 1$,
and equate it with the left side:
\begin{equation*}
	\frac{\mathrm d \subsample}{\mathrm d n} = - \frac{\subexp \subsample(n)}{n}
\end{equation*}
Integrating:
\begin{equation*}
	\subsample(n)=\submultcoef n^{-\subexp}
\end{equation*}
The parameter, 
$\submultcoef$, is set such that $n_1^\prime=n_1$:
\begin{equation*}
	C = n_1^\subexp
\end{equation*}
while $\subexp$ is set such that:
\begin{eqnarray*}
	\datafraction \sum_i n_i & = & \sum_i \subsample(n_i) n_i \\
			      & = & n_1^\subexp \sum_i n_i^{1-\subexp}
\end{eqnarray*}
where $0 < \datafraction < 1$ is the desired fraction of training data.

\else
  
\fi

\newpage

\addcontentsline{toc}{section}{References}
\ifsubmit
  \bibliography{agf_bib,svm_accel,pwl,datasets}
\else
  \bibliography{../../agf_bib,../svm_accel,../../pwl,../../datasets.bib}
\fi

\end{document}